\documentclass[sigconf,nonacm]{acmart}

\usepackage{amsmath}
\usepackage{array}
\usepackage{booktabs}
\usepackage{graphicx}
\usepackage{microtype}
\usepackage{xcolor}

\setcopyright{none}
\renewcommand\footnotetextcopyrightpermission[1]{}
\acmConference[KDD Eval Workshop '26]{KDD Workshop on Evaluation and Trustworthiness of Agentic AI}{August 9, 2026}{Jeju, Korea}
\acmYear{2026}
\copyrightyear{2026}
\acmDOI{}
\acmISBN{}

\hypersetup{
  pdfauthor={Wei-Jung Huang}
}
\newcolumntype{L}[1]{>{\raggedright\arraybackslash}p{#1}}

\title[Agent Benchmark Decisions]{\texorpdfstring{How Many Tasks Are Enough for Agent Benchmark Decisions?\\A Replay Analysis of Public LLM Agent Benchmarks}{How Many Tasks Are Enough for Agent Benchmark Decisions? A Replay Analysis of Public LLM Agent Benchmarks}}
\author{Wei-Jung Huang}
\affiliation{%
  \institution{Independent Researcher}
  \country{United States}
}
\email{william.wj.huang@gmail.com}

\begin{document}

\begin{abstract}
Agent benchmarks often compare two agents after all tasks have run, but costly evaluations make partial runs tempting. A task fraction alone does not show whether a partial run supports the same pairwise conclusion as the completed benchmark. We study this question by replaying completed public task-level records from SWE-bench, AppWorld, and tau-bench. A partial budget counts as enough only when it supports the completed benchmark's decision, covers required task groups, and leaves no more than a target fraction of comparisons unresolved. The required task fraction varies sharply. At the strict 0 percentage point threshold on a 5 percentage point budget grid, AppWorld first meets all targets at 15 percent, tau-bench at 25 percent, and SWE-bench Verified at 90 percent; SWE-bench Lite does not meet all targets by 95 percent under the primary coverage rule. Partial-evaluation reports should state how much one agent must outperform another, how tasks are selected, what coverage rule is required, what decision rule is used, and how many comparisons may remain unresolved.
\end{abstract}



\maketitle

\section{Introduction}

Agent benchmarks turn many task outcomes into comparisons between systems. In this paper, a system is a public leaderboard entry, including the model, prompt, tools, scaffolding, retry logic, and other components that produce task outcomes. Once every task has run, the benchmark gives each system a success rate. For a pair of systems, a benchmark user may then ask whether system \(A\)'s paired success rate is high enough relative to system \(B\)'s to justify preferring \(A\) for the decision at hand.

The improvement threshold should match that decision. A 0 percentage point (0 pp) threshold measures direction only: whether \(A\)'s paired success rate is strictly higher than \(B\)'s. Positive thresholds, such as 5 pp or 10 pp, require a larger margin before preferring one system over another. In all cases, the threshold defines the full-benchmark decision made after all tasks have run.

Running fewer tasks can serve two goals. One goal is score estimation: if the completed benchmark would give a system a 55 percent success rate, can a partial run estimate that number accurately enough? This paper studies a different goal: if the completed benchmark would conclude that \(A\) is better than \(B\), can a partial run support the same conclusion under stated error targets, coverage rules, and a limit on cases with no partial conclusion? A score is a measurement. A pairwise conclusion is a decision.

A partial evaluation observes only the tasks selected within its budget. It may see a fixed fraction of tasks, then return a conclusion, spend more budget, or stop without a conclusion because the observed tasks are not enough. We call such cases unresolved comparisons. In the technical notation, the same outcome is denoted as deferral. For example, if the threshold is 5 pp and \(A\) is ahead by 4 tasks among 100 observed tasks, the partial run has produced a score, but it has not shown that \(A\) clears the threshold. If uncertainty in the observed subset also prevents the evaluator from concluding that \(A\) does not clear the threshold, the rule leaves the comparison unresolved rather than forcing a conclusion.

That distinction matters because partial evaluation can fail in ways that score estimation does not capture. A partial score can be close to the final score while missing a task group that changes the pairwise conclusion. A low-error partial rule can also look reliable because it decides only easy comparisons and leaves harder ones unresolved. A cheap-first ordering can save cost while observing tasks that are not informative for the pair being compared. These are decision failures, not just estimation errors.

A small example makes the problem concrete. Suppose a partial run observes only tasks from one repository, where \(A\) beats \(B\). If the unobserved repositories favor \(B\) strongly enough, the completed benchmark can reach the opposite pairwise conclusion. A report that states only the partial task fraction or the partial score would not say whether the observed tasks covered the groups needed for the comparison. The same issue appears when a rule withholds many difficult comparisons: its reported error rate can be low because many hard cases never receive a partial conclusion.

We therefore analyze completed benchmark records by replaying partial task runs. The task set, systems, and task outcomes are fixed; what varies is which tasks have been observed before the budget is exhausted. The target is the full-benchmark decision induced by the completed record. Claims about future tasks, population-level sampling, or formal online stopping require a different model.

The resulting question is narrower than score estimation or online stopping: given a threshold, a coverage rule, a target on unresolved comparisons, a rule for choosing tasks, and a rule for when the partial run may conclude, how many tasks are needed before the partial run matches the full-benchmark decision often enough for the stated use case?

The central quantity is the minimum sufficient task budget: the smallest tested budget where the partial run is reliable enough for the stated decision. To estimate it, the analysis starts from completed task-level records and replays partial task orders. For each budgeted rule, it asks whether partial conclusions match the completed benchmark, whether the observed tasks cover the required task groups, and how often the rule lacks enough evidence for a final conclusion. A task budget is sufficient only when decision-error targets, task-group coverage targets, and unresolved-comparison targets are all met.

The empirical evidence is public, and each source plays a different role. SWE-bench Lite and Verified provide task-level software-repair outcomes for many public leaderboard entries \cite{jimenez2024swebench,swebenchleaderboard}. SWE-bench Lite is the large case where budget is measured by number of tasks. SWE-bench Verified is the case where low replay error can still leave too many comparisons unresolved. AppWorld provides task-level outcomes for interactive coding and app-use agents \cite{trivedi2024appworld}, which tests whether the same partial-evaluation question applies outside software repair. The tau-bench public trajectory releases provide a grid with cost and reward-handling fields \cite{yao2024taubench,barres2025tau2}, which lets us test one cheap-first failure mode.

\textbf{Main findings.} Public records show why task fraction alone is not a decision rule. At the strict 0 pp threshold on a 5 percentage point budget grid, AppWorld reaches sufficiency at 15 percent, tau-bench at 25 percent, and SWE-bench Verified at 90 percent. SWE-bench Lite does not reach sufficiency by 95 percent under the primary coverage rule. In that case, early budgets can control decision error and coverage failure, but too many comparisons remain unresolved.

The exact budget depends on the threshold, coverage rule, pair orientation, and task ordering. At 5 pp and 10 pp, SWE-bench Lite reaches sufficiency earlier than it does at 0 pp. With a one-task slack version of the coverage rule, the 0 pp SWE-bench Lite budget becomes sufficient at 75 percent on the 25/50/75 percent grid. Keeping only one orientation per unordered system pair is a dependence check rather than a replacement for the directed comparison; it can change positive and negative support and some sufficient-budget calls, so we report pair orientation explicitly.

The paper makes three contributions:
\begin{enumerate}
    \item It identifies the reporting failure: a partial task fraction or partial score can hide wrong pairwise decisions, missing task groups, or too many unresolved comparisons.
    \item It defines the minimum sufficient task budget, the smallest tested budget that satisfies decision-error targets, task-group coverage targets, and unresolved-comparison targets for a stated threshold.
    \item It measures this quantity on public SWE-bench, AppWorld, and tau-bench records, showing that the answer changes with benchmark, threshold, coverage rule, target on unresolved comparisons, pair orientation, and task ordering.
\end{enumerate}

\section{Related Work}

\subsection*{Estimating Scores With Less Evaluation}
Efficient evaluation methods usually address score estimation: how accurately can a benchmark score be estimated from fewer items? HELM standardizes scenarios and metrics for broad language-model evaluation \cite{liang2023helm}. MT-Bench with Chatbot Arena standardizes pairwise preference collection for chat assistants \cite{zheng2023judging}. Cost-efficient LLM evaluation asks how many examples, queries, or benchmark items are needed to estimate model performance.

Several statistical tools fit that score-estimation goal. Cer-Eval frames test-data sufficiency through confidence intervals for LLM evaluation \cite{wang2025cereval}. FAQ gives confidence intervals for performance estimates under a finite population view of benchmark evaluation \cite{wu2026faq}. Finite-population inequalities for sampling without replacement and stratified allocation give tools for partial benchmark use \cite{serfling1974probability,neyman1934representative}.

Those methods are useful when the target is a mean score or a confidence interval. Our target is whether a partial task set supports the same pairwise decision as the completed benchmark. A score interval can inform that decision, but it does not specify the whole decision policy: what threshold matters, what task groups must be covered, which decision errors are acceptable, or how often the evaluator may withhold a final decision. We therefore report decision errors, task-group coverage, and unresolved comparisons in addition to partial scores.

\subsection*{Agent Benchmarks and Cost}
Agent benchmarks make the decision problem concrete because their tasks are expensive, heterogeneous, and tied to downstream use. Aggregate success rates can hide cost, reproducibility, and reliability issues. Kapoor et al.~\cite{kapoor2024agentsmatter} argue that agent evaluation should account for accuracy, cost, downstream developer needs, overfitting, and reproducibility. SWE-bench evaluates software agents on real GitHub issue resolution \cite{jimenez2024swebench}. AppWorld evaluates interactive coding agents that operate apps through APIs with programmatic state-based checks \cite{trivedi2024appworld}. tau-bench and tau2-bench evaluate tool use in agent-user interaction in realistic domains \cite{yao2024taubench,barres2025tau2}; the public trajectory files used here also include cost fields.

Our analysis starts after these benchmarks have fixed their tasks, systems, and pass/fail outcomes. At that point the evaluator still needs a decision rule for a pair of systems under a partial budget. The same partial task fraction can support a decision, fail coverage, or save cost while observing the wrong task groups.

\subsection*{Stopping Rules and Deferral}
Sequential testing and confidence sequences provide tools for stopping under uncertainty \cite{wald1945sequential,howard2021confidence}. Selective prediction and learning to defer address settings in which a system withholds a prediction or routes a decision to another decision maker \cite{geifman2019selectivenet,mozannar2020defer}. Conformal risk control studies calibrated risk constraints under explicit assumptions \cite{angelopoulos2022crc}.

Our use of deferral is narrower. In this paper, deferral means that a partial policy leaves the comparison unresolved at the budget. Because a completed benchmark record is replayed under partial observation, the measurement target is not online stopping or population risk control. It is whether a partial run preserves the completed benchmark's pairwise decision under stated coverage and targets on unresolved comparisons.

\section{Methodology}

\subsection*{From Partial Scores to Decisions}

The setup starts from a completed benchmark record. Two systems have pass/fail outcomes on the same tasks. The completed record tells us how much the tested system outperformed the comparison system on that fixed task set. Once the evaluator chooses an improvement threshold, the completed benchmark induces the full-benchmark decision: a positive conclusion for the tested system if its paired improvement is strictly above the threshold, and a negative conclusion otherwise. Metric names below use false accept and false reject for these two error directions. Ordinary prose uses positive conclusion, negative conclusion, and unresolved comparison.

The replay asks what would have happened with less evidence. If the evaluator had seen only a fixed fraction of the task set, would the observed subset have led to the same pairwise conclusion? A useful partial evaluation must do more than match the final decision on easy cases. It must cover the relevant task groups and return final conclusions often enough to be useful.

Formally, let \(T=\{t_1,\ldots,t_N\}\) be the fixed task set. Each task can have metadata such as repository, domain, difficulty, or expected cost. For an ordered pairwise comparison, let \(A\) be the tested system and \(B\) be the comparison system. The comparison asks a one-sided evaluation question: does tested system \(A\) clear threshold \(\delta\) over comparison system \(B\)? The two orientations of an unordered pair share the same task outcomes. The ordered count therefore counts evaluation questions, while the unordered count describes system-pair evidence. We report both counts and also repeat the analysis after keeping one orientation per unordered pair.

For task \(t_i\), \(Y_{i,A}\in\{0,1\}\) and \(Y_{i,B}\in\{0,1\}\) are verified outcomes. The paired task difference is
\begin{equation}
    D_i = Y_{i,A} - Y_{i,B}.
\end{equation}
The completed task set gives
\begin{equation}
    \Delta_{\mathrm{full}} = \frac{1}{N}\sum_{i=1}^{N} D_i.
\end{equation}
For a subset \(S_b\) observed by budget \(b\), the partial paired improvement is
\begin{equation}
    \widehat{\Delta}_{b} = \frac{1}{|S_b|}\sum_{i:t_i\in S_b} D_i.
\end{equation}

\paragraph{Threshold.}
Before replaying partial evaluations, the evaluator sets an improvement threshold \(\delta\). The evaluator supplies this threshold as part of the decision problem; the benchmark supplies the task outcomes. The threshold can be zero when the question is only whether one system scores strictly above another, or positive when the evaluator requires a larger margin before preferring the tested system. The full-benchmark decision is positive for the tested system when \(\Delta_{\mathrm{full}} > \delta\) and negative otherwise. This is the decision that each replay tries to match.

The threshold should be set before replay to the smallest improvement that would justify preferring the tested system over the comparison system, given evaluation cost and operational risk. The public experiments report 0 pp, 5 pp, and 10 pp. The 0 pp setting measures strict ordering, where one-task differences can decide the full-benchmark label. The 5 pp and 10 pp settings show how the sufficient task budget changes when the evaluator requires a larger improvement.

The analysis can be rerun at any chosen threshold, so a practitioner should use the threshold that matches the decision they face. We report the number of positive and negative full-benchmark decisions at each threshold because larger thresholds can change the problem itself: they can make positive decisions rare, which can make some error rates look easier to satisfy.

\paragraph{Policy outputs.}
A budgeted policy observes a subset \(S_b \subset T\) before reaching budget \(b\). Along a budget schedule, the rule may continue sampling at intermediate budgets. At a reported budget cap, it returns a positive conclusion for \(A\), returns a negative conclusion for \(A\), or leaves the comparison unresolved. The equations denote these three outputs as accept, reject, and defer, but the prose uses the more descriptive terms above.

An unresolved comparison means the budget is exhausted and the observed tasks do not support either final decision. The rule withholds a conclusion by design: a partial report should not convert weak evidence into a final pairwise claim. At a 5 pp threshold, for instance, a 4 pp partial advantage for \(A\) does not support a positive conclusion. If uncertainty in the observed subset also prevents a confident negative conclusion, the comparison remains unresolved.

\paragraph{Coverage.}
Coverage is designed to prevent a specific error: a partial run can see a narrow task slice, reach a confident pairwise conclusion within that slice, and miss groups that would change the completed-benchmark decision. The relevant groups come from metadata available in each source: repository for SWE-bench, split by difficulty for AppWorld, and domain for tau-bench. The primary rule therefore requires every group to appear, with larger groups represented roughly in proportion to their share of the fixed task set.

Let \(q_i\) be the group for task \(t_i\), \(n_g\) the number of tasks in group \(g\), \(b\in(0,1]\) the task-budget fraction, and \(k_b=\lceil bN\rceil\). The required count for group \(g\) is
\begin{equation}
    r_g(b)=\max\{1,\min\{n_g,\lfloor k_b n_g/N\rfloor\}\}.
\end{equation}

An observed subset satisfies coverage when it contains at least \(r_g(b)\) tasks from every group \(g\). This floor approximates the completed benchmark's group mixture at budget \(b\) while guarding against the loss of a task group that appears in the completed benchmark. If the sum of the required counts exceeds the current budget, then coverage is infeasible at that budget and the coverage-aware rule leaves the comparison unresolved. A coverage failure occurs when a policy returns a final conclusion before satisfying these group-count requirements.

The slack floor tests how much this proportional requirement affects the reported budget. It uses
\begin{equation}
    r^{\mathrm{slack}}_g(b)=\max\{1,\min\{n_g,\lfloor k_b n_g/N\rfloor - 1\}\},
\end{equation}
which subtracts one task from the proportional floor while still requiring every group to appear.

\paragraph{Replay outcomes.}
We measure four outcomes against the full-benchmark decision. A false accept occurs when a partial policy gives a positive conclusion for \(A\) when the full benchmark gives a negative conclusion. A false reject occurs when a partial policy gives a negative conclusion for \(A\) when the full benchmark gives a positive conclusion. Coverage failure records final conclusions made before the observed tasks cover the required task groups. The unresolved-comparison rate records cases where the policy leaves the comparison unresolved within budget.

Together, these quantities define what the replay measures for a completed benchmark record. For a fixed completed record, threshold, budget, and policy, we average the policy's output over sampled task orders and ordered system pairs.

\paragraph{Error denominators.}
Conditional false-accept and false-reject rates are conditional on the full-benchmark decision class, not on whether the policy returns a final conclusion. The conditional false-accept denominator is all replayed cases whose full-benchmark decision is negative for \(A\); unresolved cases stay in that denominator and count as neither false accepts nor correct negative conclusions. The conditional false-reject denominator is all replayed cases whose full-benchmark decision is positive for \(A\); unresolved cases again stay in the denominator.

Coverage failure and rates of unresolved comparisons are averaged over all replayed cases. The result describes the behavior of a chosen partial-evaluation policy on the completed benchmark record, not an estimate of future benchmark performance.

\subsection*{When a Budget Counts as Sufficient}

A partial budget counts as sufficient only when it rules out three ways a partial run can mislead. First, the policy should rarely give the wrong positive or negative conclusion relative to the completed benchmark. Second, it should not give a final conclusion before the observed tasks cover the required groups. Third, it should resolve enough comparisons to be useful, rather than achieving low error by declining hard comparisons. The target on unresolved comparisons sets the largest fraction of replayed cases that may end without a partial conclusion at the budget.

We define the minimum sufficient task budget as the smallest tested task budget \(b\) for which
\begin{align}
    \mathrm{FAR}_{\mathrm{cond}}(b) &\leq \alpha,\\
    \mathrm{FRR}_{\mathrm{cond}}(b) &\leq \alpha,\\
    \mathrm{coverage\ failure}(b) &\leq \gamma,\\
    \mathrm{defer}(b) &\leq \eta.
\end{align}
Here \(\mathrm{FAR}_{\mathrm{cond}}\) is measured over negative full-benchmark cases. \(\mathrm{FRR}_{\mathrm{cond}}\) is measured over positive full-benchmark cases. We also report the smallest task budget that meets the error and coverage targets without constraining unresolved comparisons.

The fourth line is the rate of unresolved comparisons, denoted \(\mathrm{defer}(b)\) in the equations. This line prevents a policy that resolves only easy cases from appearing reliable merely because its false-accept and false-reject rates are low. Unless otherwise stated, \(\alpha=0.05\), \(\gamma=0.05\), and \(\eta=0.25\). Together, these targets define a fixed comparison point across public sources: 5 percent conditional decision-error and coverage-failure targets, plus a requirement that at most 25 percent of comparisons remain unresolved. A practitioner should set these targets before replay to match the risk and cost of the evaluation decision.

The threshold \(\delta\) and unresolved-comparison target \(\eta\) have different roles. The threshold states how large a paired improvement would justify preferring \(A\) over \(B\) after the completed benchmark. The unresolved-comparison target states how often a partial report may return no conclusion. In a leaderboard triage setting, a larger \(\eta\) may be acceptable if unresolved pairs are simply sent to the full benchmark. In a deployment or procurement setting, a smaller \(\eta\) may be needed because the partial report is expected to resolve most comparisons. The public experiments therefore report 0 pp, 5 pp, and 10 pp thresholds under a fixed \(\eta=0.25\), but these values are examples of the replay report rather than universal evaluation rules.

The replay target is whether the partial task run matches the full-benchmark decision over different task orders in a fixed, completed task set. Because those orders are permutations of completed records, the reported rates describe those records. All sufficiency claims in this paper are empirical replay claims under stated thresholds, policies, and targets.

\begin{figure*}[t]
    \centering
    \includegraphics[width=0.95\textwidth]{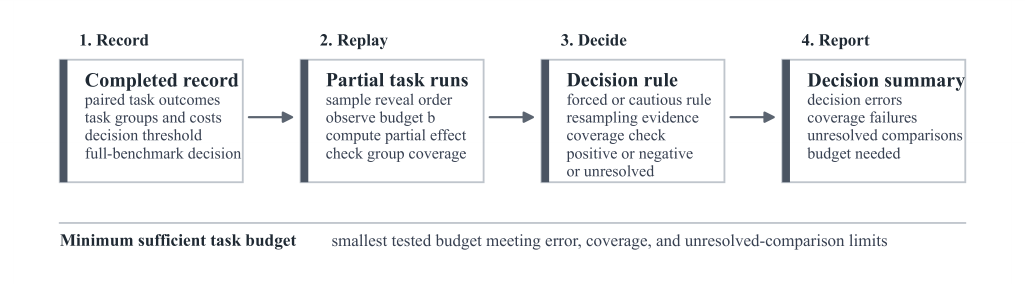}
    \Description{Flow diagram showing completed benchmark records, partial replay, and a decision report with decision error, task-group coverage, unresolved comparisons, and minimum budget.}
    \caption{Replay tests whether partial task evidence can support the same pairwise decision as the completed benchmark.}
    \label{fig:decision-layer}
\end{figure*}

Figure~\ref{fig:decision-layer} summarizes this workflow. Completed task-level records are replayed under partial budgets to test whether a partial run can support the same pairwise decision as the completed benchmark. Each replayed case records decision errors, coverage failures, and unresolved comparisons. The aggregate report then records the minimum tested budget that meets all targets.

\subsection*{Policies Compared in Replay}

The replay analysis uses completed records to simulate many partial evaluations without changing task outcomes. For each ordered pairwise comparison, we sample reveal orders over tasks. At each budget point, a policy observes the tasks that fall within the budget, computes the partial paired effect, checks task-group coverage, and either returns a final decision or leaves the comparison unresolved. The replay records false accept, false reject, overall decision error, unresolved comparisons, coverage failure, cost fraction when available, and budget used.

\paragraph{Task-ordering policies.}
The same replay procedure can evaluate any rule that chooses tasks and maps observed outcomes to a final conclusion, a request for more tasks, or an unresolved comparison. We focus on policy families that represent practical evaluation choices.

Forced policies must return a final conclusion at the budget, so they show what can go wrong when an evaluator insists on a decision. Deferring policies may leave the comparison unresolved when the observed tasks are not enough for either final decision.

Uniform forced policies sample tasks without group or cost metadata. Group-stratified forced policies spread task groups across the reveal order, using repository for SWE-bench, split by difficulty for AppWorld, and domain for tau-bench. Cost-aware forced policies, available only when cost fields exist, reveal tasks in increasing tested-system cost.

\paragraph{Bootstrap policies.}
Bootstrap tail policies are the main policies that can leave comparisons unresolved. They work directly with paired binary task outcomes, give a tunable uncertainty check, and withhold a final decision when the observed subset is too close to the threshold. They make a final decision only when most resamples agree that the observed paired improvement is on the same side of the threshold as \(\widehat{\Delta}_b\). The coverage-aware version also requires task-group coverage before making a final decision.

For coverage-aware bootstrap, write \(a\) for the bootstrap tail-probability cutoff used by the policy. Larger \(a\) values make the policy less cautious, because fewer resamples must stay away from the opposite side of the threshold. The policy returns a positive conclusion only if the sample mean is strictly above the threshold and at most a fraction \(a\) of bootstrap resamples are at or below the threshold. It returns a negative conclusion only if the sample mean is at or below the threshold and at most a fraction \(a\) of bootstrap resamples are above the threshold. Otherwise, the comparison remains unresolved.

The main policy uses \(a=0.05\). This has the same numerical value as the replay-level conditional error target \(\alpha\), but the two quantities have different roles: \(a\) is a policy tuning parameter, while \(\alpha\) is an empirical replay target. The empirical replay determines whether false accepts, false rejects, coverage failures, and unresolved comparisons are all within their stated limits.

Bootstrap tail quantities summarize the observed subset; they are not population guarantees. The completed task set is fixed, and the replay measures how decisions change with the tasks observed at a budget. We therefore report a sweep over \(a\in\{0.02,0.05,0.10\}\) and a sweep over bootstrap replicate counts.

\paragraph{Comparator policies.}
Two comparator families separate decision-rule effects from task-allocation effects. Statistical decision-rule comparators test whether the bootstrap decision rule itself drives the budget. A coverage paired-normal test applies a one-sided paired mean test to the observed paired differences against threshold \(\delta\). At the 0 pp threshold, where binary disagreements directly define the directional comparison, we also report a coverage exact McNemar test.

Second, allocation comparators test whether different task allocation could reduce the budget. A Neyman coverage bootstrap first satisfies the group-coverage floor, then allocates the remaining task budget in proportion to \(N_g S_g\), where \(N_g\) is the group size and \(S_g\) is the completed-record standard deviation of paired differences in group \(g\). This diagnostic estimates how much variance information could help; a live evaluator requires pilot or historical records to estimate those variances before evaluation. A Serfling coverage policy uses a finite-population Hoeffding-Serfling radius on the same stratified task subsets. Together, these comparisons test whether budgets are mainly driven by allocation, uncertainty in the observed task subsets, or the bootstrap decision rule.

The reveal policies fix task order before observing the current partial effect. They cover practical choices an evaluator might make: run a random subset, enforce task-group coverage, prefer cheaper tasks, require resampling evidence, or require both resampling evidence and task-group coverage. Comparing them separates effects that a single aggregate score hides: wrong decisions, unresolved comparisons, task coverage, and cost.

\section{Experiments}

\subsection*{Public Records}

The public sources cover complementary parts of the decision question. SWE-bench Lite is the large task-count case, where repository coverage, unresolved comparisons, and threshold choice can all change the budget. SWE-bench Verified tests whether low decision error can still leave too many comparisons unresolved. AppWorld asks whether the same partial-budget issue appears outside software repair. The tau-bench public trajectory release supplies the cost-ordered check, where a cheap-first task order can save cost while missing the domains needed for the completed-benchmark decision.

Table~\ref{tab:evidence-sources} summarizes the records and the role each source plays in the evidence. Pair counts are ordered pairwise comparisons, with unordered pair counts in parentheses. Positive counts are ordered full-benchmark cases where the tested system clears the threshold at 0 pp, 5 pp, and 10 pp, using the same strict greater-than decision rule as the replay. These counts matter because the threshold changes the number of positive and negative cases available for measuring conditional decision errors. SWE-bench Lite and Verified use budgets measured by task count with repository grouping. AppWorld uses a task-count budget outside software repair, with split-by-difficulty grouping. The tau-bench public grid uses available cost fields and domain grouping.

\begin{table*}[t]
    \centering
    \small
    \caption{Public records used in the replay. Pair counts are ordered, with unordered pair counts in parentheses.}
    \label{tab:evidence-sources}
    \scriptsize
    \begin{tabular}{L{2.4cm}rrrrrL{2.0cm}L{1.0cm}L{2.7cm}}
        \toprule
        Source & Pairs & Task units & Positive 0 pp & Positive 5 pp & Positive 10 pp & Grouping & Cost & Evidence role \\
        \midrule
        SWE-bench Lite & 552 (276) & 300 median overlap & 275 & 199 & 154 & repository & no & large task-count case \\
        SWE-bench Verified & 156 (78) & 496 median overlap & 77 & 60 & 54 & repository & no & many unresolved cases \\
        AppWorld leaderboard & 306 (153) & 585 tasks & 153 & 128 & 108 & split-difficulty & no & outside software repair \\
        tau-bench public grid & 12 (6) & 278 domain-task items & 6 & 5 & 2 & domain & yes & cheap-first cost check \\
        \bottomrule
    \end{tabular}
\end{table*}

For SWE-bench, we use public result metadata from the experiments repository after filtering to near-complete leaderboard entries. Near-complete means that a public entry reports outcomes for at least 95 percent of the split. The Lite split provides 552 ordered pairwise comparisons from 276 unordered system pairs, with median overlap of 300 tasks. The Verified split provides 156 ordered pairwise comparisons from 78 unordered pairs, with median overlap of 496 tasks. The metadata used here lack a cost field, so these are task-count replays.

For AppWorld, we unpacked all 36 public leaderboard bundles with the official AppWorld tooling and parsed task-level \texttt{success} labels from the released evaluation files. The 36 bundles correspond to two split bundles for each of 18 systems. Combining the two splits yields 585 task keys and 306 ordered pairwise comparisons from 153 unordered system pairs. AppWorld tests the same partial-budget question outside software repair while still providing public task-level labels, stable task IDs, difficulty labels, and scenario IDs.

For the tau-bench public trajectory grid, the unit of analysis is a domain-task item. The public grid uses four leaderboard entries, all 12 ordered pairwise comparisons among them, 278 domain-task items per system, and three text domains: airline, retail, and telecom. Banking-knowledge trajectories are excluded because retrieval configuration differs across the selected entries. The selected public files provide the cost and reward-handling fields needed for a focused cost-ordering check: when tasks are sorted by tested-system cost, does the partial run still cover the domains needed for the completed-benchmark decision?

\subsection*{Replay Settings and Reproducibility}

\paragraph{Replay sampling.}
For SWE-bench and AppWorld, we use 500 reveal orders per ordered pairwise comparison and 200 bootstrap samples per policy evaluation. For tau-bench, we use 2{,}000 reveal orders and 500 bootstrap samples. At each budget, replay metrics are averaged over ordered system pairs and sampled reveal orders.

\paragraph{Targets and grids.}
A task budget must meet a 5 percent conditional decision-error target, a 5 percent coverage-failure target, and a maximum unresolved-comparison rate of 25 percent. This common setting keeps the public records comparable across benchmarks. The main budget search uses a 5 percentage point grid from 5 to 95 percent. We also report the coarser 25, 50, and 75 percent grid because many partial-evaluation reports use these round budgets.

The 0 pp setting asks whether the tested system scores strictly above the comparison system on the completed task set. In this setting, one-task differences can be enough to change the full-benchmark decision. The 5 pp and 10 pp settings show how the sufficient task budget changes when the evaluator requires a larger improvement before preferring one system over another.

\paragraph{Reproducibility package.}
The public package is available at \url{https://github.com/WilliamWJHuang/How-Many-Tasks-Are-Enough-for-Agent-Benchmark-Decisions}. It contains replay code, schemas, manifests, fixed seeds, and derived tables for regenerating the SWE-bench, AppWorld, and tau-bench summaries. It also contains checks for the bootstrap tail-probability cutoff, Neyman and Serfling comparison policies, reveal-order seed intervals, one-orientation-per-pair analysis, grouping and coverage floors, and tau-bench cost resampling. Raw AppWorld leaderboard bundles remain in the public leaderboard repository, and the package instructions unpack them with the official tooling.

\section{Results}

The main empirical question is how much of a completed benchmark must be observed before the partial run supports the same pairwise conclusion. We first report the minimum sufficient task budgets under the primary coverage-aware bootstrap policy. We then use policy, threshold, grouping, orientation, and cost checks to show why the task fraction alone is not enough to interpret a partial evaluation.

\begin{figure*}[t]
    \centering
    \includegraphics[width=0.95\textwidth]{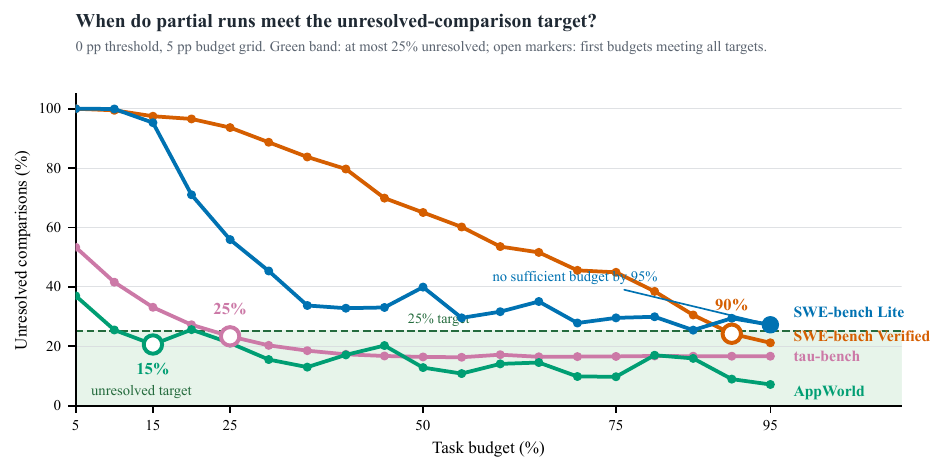}
    \Description{Line chart for the 0 percentage point threshold showing unresolved comparisons as task budget increases from 5 to 95 percent. AppWorld reaches its first sufficient budget at 15 percent, tau-bench at 25 percent, SWE-bench Verified at 90 percent, and SWE-bench Lite has no sufficient budget by 95 percent.}
    \caption{Unresolved comparisons across the 5 percentage point budget grid at the 0 pp threshold. The green band marks the unresolved-comparison target; labeled open markers denote the first budgets that also meet the decision-error and coverage targets.}
    \label{fig:public-deferral-curve}
\end{figure*}

Figure~\ref{fig:public-deferral-curve} gives the main empirical pattern at the 0 pp threshold on the 5 percentage point budget grid. The vertical axis is the percentage of replayed comparisons that remain unresolved at a given task budget. The green band marks the unresolved-comparison target, while the labeled open markers require all targets to be met: decision error, task-group coverage, and unresolved comparisons. AppWorld and tau-bench reach sufficiency early, SWE-bench Verified reaches it only near the end of the grid, and SWE-bench Lite does not reach it by 95 percent. The 5 pp and 10 pp thresholds, coverage floors, task-ordering policies, and sensitivity checks then show which design choices move these budgets.

\subsection*{Task-Count Budget Replays}

\paragraph{How much budget is enough?}
There is no single task fraction that is enough. At 0 pp, the minimum sufficient task budget ranges from 15 percent for AppWorld to 90 percent for SWE-bench Verified on the 5 percentage point grid. tau-bench reaches sufficiency at 25 percent, while SWE-bench Lite does not reach sufficiency by 95 percent under the primary coverage rule. The main reason is not only wrong decisions; some budgets avoid most wrong decisions but still leave too many comparisons without a partial conclusion.

\paragraph{What changes with the threshold?}
Threshold choice changes the answer because it changes the full-benchmark decision being replayed. AppWorld remains sufficient at 15 percent for all three thresholds, and SWE-bench Verified remains a high-budget case at 90 percent. tau-bench moves from 25 percent at 0 pp to 35 percent at 10 pp. SWE-bench Lite changes more sharply: it has no sufficient budget by 95 percent at 0 pp, but reaches 55 percent at 5 pp and 35 percent at 10 pp.

\paragraph{Why does SWE-bench Lite fail at 0 pp?}
SWE-bench Lite shows why an error-only report would be misleading. At 0 pp, the primary bootstrap policy meets the error-and-coverage targets at 25 percent task budget, but 55.92 percent of comparisons are still unresolved. Even at 95 percent task budget, 27.25 percent remain unresolved, which is above the 25 percent target. Thus, no tested budget on the 5 percentage point grid is sufficient at 0 pp, even though the policy's final decisions are usually correct when it makes them.

Positive-margin thresholds and coverage floors change the SWE-bench Lite conclusion. With the same primary coverage rule, the sufficient budget is 55 percent at 5 pp and 35 percent at 10 pp. With a one-task slack coverage rule on the 25/50/75 percent grid, the 0 pp case becomes sufficient at 75 percent. The lesson is that a reported task budget is meaningful only together with the threshold and coverage rule that produced it.

\paragraph{Why does SWE-bench Verified need late budgets?}
SWE-bench Verified makes the unresolved-comparison target more visible. At 0 pp, the primary policy meets the error-and-coverage targets at 25 percent task budget, but leaves 93.64 percent of comparisons unresolved. The same policy does not meet the unresolved-comparison target until 90 percent task budget, where 24.22 percent of comparisons remain unresolved. At 5 pp and 10 pp, SWE-bench Verified also first reaches sufficiency at 90 percent.

\paragraph{Why does AppWorld pass early?}
AppWorld is the case where early partial evaluation works under the stated targets. The primary policy reaches sufficiency at 15 percent for 0 pp, 5 pp, and 10 pp, with about 21 percent unresolved comparisons at the sufficient budget. The policy comparison shows why the task fraction itself is not the explanation. Split-difficulty stratified forced evaluation needs 50 percent task budget, while uniform forced evaluation fails the coverage target at 25 percent with 99.96 percent coverage failure. The same fraction can therefore support a final decision under one task-ordering rule and fail under another.

\subsection*{tau-bench Cost Check}

tau-bench tests cheap-first ordering: does running cheaper tasks first preserve the decision? In this public grid, coverage-aware bootstrap reaches sufficiency at 25 percent for 0 pp, 30 percent for 5 pp, and 35 percent for 10 pp. The role of this source is focused: the selected public trajectory files include both domain labels and cost fields, so they test whether a low-cost reveal order also observes the domains needed for the completed-benchmark decision.

Cost ordering shows the failure mode directly. At 25 percent task budget, cost-aware forced evaluation uses only 11.51 percent of tested-system cost on average, but has 100 percent coverage failure. It also produces wrong pairwise conclusions at all three thresholds, including false rejections in 4 of 5 positive 5 pp cases and 2 of 2 positive 10 pp cases. Cheap-first task orders can save cost while missing the task groups needed to match the full-benchmark decision.

\subsection*{What Partial Scores Hide}

Across sources, the replay exposes three failure modes that a partial score hides. First, forced partial decisions can miss the required group-count coverage. At 25 percent task budget, uniform forced evaluation has 99.96 percent coverage failure on both SWE-bench Lite and AppWorld. The observed task fraction by itself cannot justify the decision.

Second, low-cost task subsets can distort the decision in the tau-bench cost check. Cost-aware forced evaluation uses only 11.51 percent of tested-system cost at 25 percent task budget, but has 100 percent coverage failure and several wrong pairwise conclusions among the 12 ordered comparisons. Low cost by itself cannot support the full-benchmark decision when the observed tasks miss required groups.

Third, low replay error can come from unresolved comparisons. On SWE-bench Verified, the bootstrap policy with repository coverage controls decision error at 25 percent task budget but leaves 93.64 percent of comparisons unresolved at 0 pp. An evaluation report should therefore state both error and the rate of unresolved comparisons.

\subsection*{Sensitivity and Comparator Checks}

The main task-budget results use the coverage-aware bootstrap policy with the primary coverage rule for each source, then report how the answer changes at 0 pp, 5 pp, and 10 pp. The checks below ask which design choices move the conclusion. Each check changes one part of the replay question while keeping the same definition of sufficiency.

\paragraph{Does the bootstrap cutoff drive the budget?}
The bootstrap cutoff controls the tradeoff between error and unresolved comparisons. A looser cutoff usually resolves more comparisons earlier, but it must still pass the replay error targets. At 0 pp, AppWorld reaches sufficiency at 10 percent with \(a=0.10\), 15 percent with \(a=0.05\), and 25 percent with \(a=0.02\). tau-bench moves in the same direction, while SWE-bench Verified remains in the high-budget regime. The cutoff is therefore part of the policy specification: changing it can change the budget even when the task set and threshold stay fixed.

\paragraph{Does the bootstrap sample count matter?}
The number of bootstrap replicates does not change which sources are early or late budget cases in these records. We rerun the primary coverage-aware bootstrap rule with 100, 200, and 500 bootstrap samples across the checked budgets and thresholds. Across the checked seeds and sample counts, the largest unresolved-comparison range for a fixed source, threshold, and budget is 1.69 percentage points. The largest conditional decision-error rate observed in the sweep is 1.40 percent, still below the 5 percent target. The sufficient-budget ordering of sources is stable over this replicate-count sweep. The reproducibility package includes the full curves.

\paragraph{Can allocation reduce the task budget?}
The allocation comparators separate task choice from the statistical decision rule. The Neyman allocation diagnostic reduces some SWE-bench Lite budgets on the 25/50/75 percent comparator grid, reaching 75 percent at 0 pp, 50 percent at 5 pp, and 25 percent at 10 pp. This diagnostic uses completed-record group variances, so it estimates how much variance information could help; a live policy requires pilot or historical variance estimates. In contrast, the Serfling finite-population policy is conservative in these records and reaches sufficiency only for AppWorld at 75 percent. Allocation information can help, but the decision rule still has to be reported.

\paragraph{Do classical paired tests change the conclusion?}
Classical paired-test rules change some budgets but not the main pattern. On the 5 percentage point budget grid, the coverage paired-normal test matches the primary bootstrap conclusion for AppWorld, tau-bench, and SWE-bench Verified at all three thresholds. On SWE-bench Lite, it still finds no sufficient 0 pp budget by 95 percent, reaches 70 percent at 5 pp, and reaches 35 percent at 10 pp.

The exact McNemar rule applies directly only at 0 pp. It is more conservative than the bootstrap rule for AppWorld and tau-bench, reaching 25 percent and 35 percent, respectively, and it moves SWE-bench Verified from 90 percent to 95 percent. The paired tests therefore change particular budgets, but the report still needs the task budget, coverage rule, and unresolved-comparison target.

\paragraph{How much do orientation and grouping matter?}
Pair orientation and grouping choices change some budgets and can change whether any tested budget is sufficient. Keeping one orientation per unordered system pair is useful as a dependence check, but it changes the support for positive and negative decisions. For example, the SWE-bench Lite undirected 0 pp collapse has 275 positive decisions and only 1 negative decision, so conditional false-accept estimates have little support. The directed orientation remains the primary target because it matches the one-sided comparison an evaluator would report.

Coverage choices matter as well. On SWE-bench Lite at 0 pp and 75 percent budget, a one-task-per-repository floor brings unresolved comparisons below the 25 percent target, while the proportional floor remains above it. A stricter repository-by-difficulty grouping moves in the other direction, leaving most Lite comparisons unresolved at the same budget. A reported task budget is therefore incomplete without the orientation and coverage rule that produced it.

\paragraph{Are the main patterns stable under replay checks?}
Reveal-order seed intervals are narrow for the reported replay metrics; for example, SWE-bench Lite's unresolved rate at 75 percent budget is 29.7 percent, with a 95 percent seed interval of 29.5 to 29.9 percent. The tau-bench cost resampling check keeps cheap-first coverage failure at 100 percent across the 5th to 95th percentile of resampling runs. SWE-bench Verified's high unresolved rate is tied to task and repository variability: moving from 75 percent to 90 percent budget raises repository coverage and brings unresolved comparisons below the target. These checks support the same interpretation: sufficiency is a property of the decision rule and task evidence, not the task fraction alone.

\section{Limitations}

The evidence supports claims about completed benchmark records. The replay estimates whether observed task subsets match the full-benchmark decision in those records. It does not estimate performance on future tasks, because the replay orders permute a fixed completed record. Extending the claim to future task draws would require an explicit sampling model and benchmark records collected under that model.

The reported budget depends on how the decision is defined. We analyze ordered evaluation questions because \(A\) versus \(B\) asks whether the tested system clears the threshold over the comparison system, while \(B\) versus \(A\) asks the reverse question. We also repeat the analysis after keeping one orientation per unordered pair, which checks how much this orientation choice changes the number of positive and negative cases and the resulting budgets. The ordered question remains the main target because it matches the one-sided comparison an evaluator would report. Thresholds, coverage floors, and grouping choices play the same role: they should be set before replay and reported with the budget.

The tested reveal policies choose random, stratified, cost-ordered, or coverage-aware task orders before observing the current partial effect. Adaptive policies from sequential testing or active task selection could reach different budgets. The Neyman comparison shows that variance-aware allocation can change sufficient budgets, but the reported version uses completed-record group variances. A live variance-aware policy requires pilot estimates, historical records, or another pre-specified source of variance information.

The grouping choices use public metadata. Repository, split by difficulty, and domain labels are available in the released records, but they need not be the only useful partitions. Learned semantic clusters or failure-mode clusters could be useful when task text, embeddings, or failure annotations are available, especially because benchmark-item correlations can affect model comparisons \cite{ailem2024benchmarkrobustness}. Adaptive coverage floors that emphasize groups most likely to change the pairwise conclusion could also reduce budgets, but they require pilot data, historical records, or another pre-specified source of group influence estimates. Such group construction would change the coverage rule and should be evaluated as another policy variable.

The public evidence covers several agent settings, but only for released task-level records. SWE-bench is software repair. AppWorld is interactive coding and app use. The tau-bench public trajectory grid is tool-agent-user interaction, and the public ordered grid used here has 12 ordered pairwise comparisons from 6 unordered pairs. Applying the same replay analysis to OSWorld, WebArena, AgentDojo, AndroidWorld, Terminal-Bench, AgentBench, or MiniWoB++ requires parsed task-level multi-system records from those benchmarks. Stronger cost-aware claims require validated per-task cost, token, latency, or interaction fields for larger public multi-system records.

\section{Conclusion}

Partial agent evaluation needs more than a cost-saving number or a partial score. The relevant question is whether the observed tasks are enough to support the same pairwise decision as the completed benchmark. We answer that question by replaying partial task runs over completed benchmark records: state the threshold, replay the partial policies, check task-group coverage, measure conditional decision errors, and report how often the policy leaves the comparison unresolved. The key quantity is the minimum sufficient task budget.

At 0 pp on the 5 percentage point budget grid, the sufficient budget ranges from 15 percent for AppWorld to 90 percent for SWE-bench Verified, while SWE-bench Lite does not reach sufficiency by 95 percent under the primary coverage rule. Positive thresholds move some budgets, especially for SWE-bench Lite and tau-bench, so the threshold remains part of the reported decision. The SWE-bench Lite slack-coverage result shows why the coverage rule must be reported with the budget. The tau-bench cost check shows that cheap-first task orders can miss the evidence needed for a decision.

The results support a reporting principle: a task fraction is not a decision rule. Evaluation reports should state whether partial evidence meets explicit targets for decision error, task coverage, and unresolved comparisons. When those targets are not met, the appropriate output is more evaluation or no partial conclusion.

\bibliographystyle{ACM-Reference-Format}
\bibliography{references}

\end{document}